\documentclass[acmsmall,screen]{acmart}
\usepackage{multirow}
\usepackage{makecell}
\usepackage{subfigure}

\AtBeginDocument{%
  }

\acmMonth{May}
\setcopyright{acmlicensed}
\copyrightyear{2024}
\acmYear{2024}
\acmVolume{23}
\acmNumber{5}

\acmDOI{10.1145/3654811}

\acmJournal{TALLIP}
\acmVolume{1}
\acmNumber{1}
\acmArticle{1}
\acmMonth{1}

\begin{document}

\title{Cleansing Jewel: A Neural Spelling Correction Model Built On Google OCR-ed Tibetan Manuscripts}

\author{Queenie Luo}
\authornote{Both authors contributed equally to this research.}
\email{queenieluo@g.harvard.edu}
\orcid{0009-0004-1854-7968}
\affiliation{%
  \institution{Harvard University}
  \streetaddress{9 Kirkland Place, 1st Floor Rear}
  \city{Cambridge}
  \state{Massachusetts}
  \country{USA}
  \postcode{02138}
}

\author{Yung-Sung Chuang}
\authornotemark[1]
\email{yungsung@mit.edu}
\orcid{0000-0002-1723-5063}
\affiliation{%
  \institution{Massachusetts Institute of Technology (MIT)}
  \streetaddress{32 Vassar Street, G-436}
  \city{Cambridge}
  \state{Massachusetts}
  \country{USA}
  \postcode{02139}
}

\renewcommand{\shortauthors}{Luo and Chuang}

\begin{abstract}
 Scholars in the humanities heavily rely on ancient manuscripts to study history, religion, and socio-political structures of the past. Significant efforts have been devoted to digitizing these precious manuscripts using OCR technology. However, most manuscripts have been blemished over the centuries, making it unrealistic for OCR programs to accurately capture faded characters. This work presents the Transformer + Confidence Score mechanism architecture for post-processing Google’s Tibetan OCR-ed outputs. According to the Loss and Character Error Rate metrics, our Transformer + Confidence Score mechanism architecture proves superior to the Transformer, LSTM-to-LSTM, and GRU-to-GRU architectures. Our method can be adapted to any language dealing with post-processing OCR outputs.
\end{abstract}

\ccsdesc[500]{Applied computing~Optical character recognition}

\keywords{Post-processing OCR output, Transformer, neural networks}

\received{20 June 2023}
\received[accepted]{17 Mar 2024}

\maketitle

\section{Introduction}\label{se:1}

Locating exact words in ancient manuscripts is a crucial but time-consuming task for scholars in the humanities. In recent years, Optical Character Recognition (OCR) technology has greatly facilitated the digitization progress of ancient texts. Buddhist Digital Resource Center (BDRC)\footnote{  BDRC's official website: \url{https://library.bdrc.io/}}  has archived over 15 million pages of culturally significant Buddhist works. Sponsored by Google's Tibetan OCR, BDRC digitized over 8,000 volumes of Tibetan Buddhist texts to create a searchable database. However, the digitized e-text corpus contains many spelling errors so it cannot be reliably used by scholars for their research. Some hand-written Tibetan graphs, such as ``\includegraphics[scale=0.019]{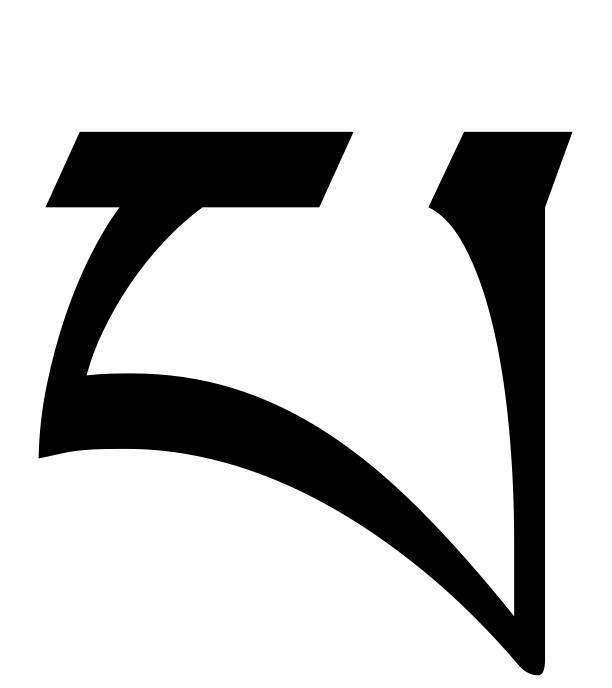}'' and ``\includegraphics[scale=0.019]{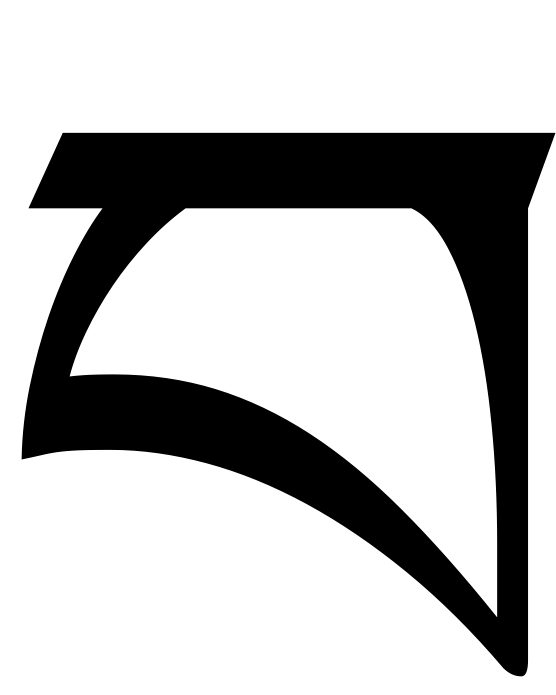},'' ``\includegraphics[scale=0.035]{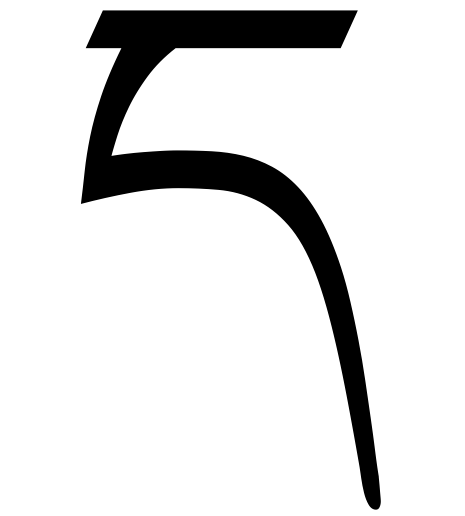}'' and ``\includegraphics[scale=0.019]{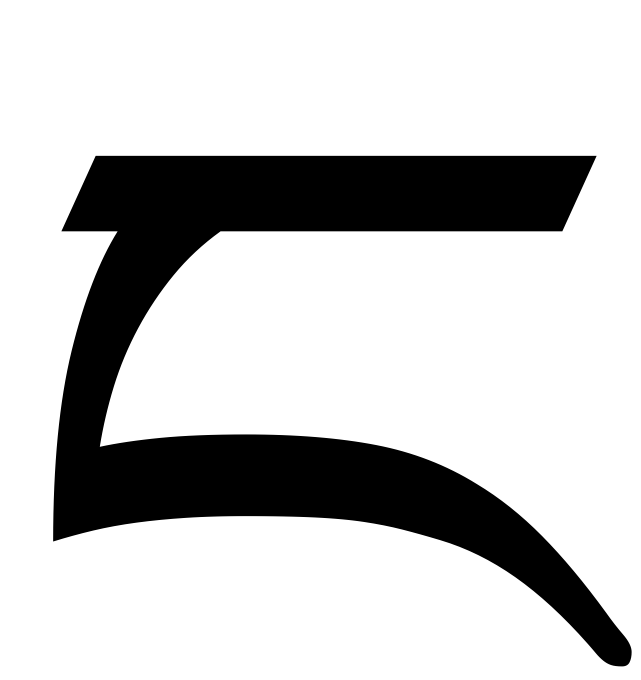},'' are often indistinguishable even to the naked eye, and an OCR program cannot be expected to capture these nuances. Scholars in this field have to rely on their linguistic knowledge to discern the correct spellings based on word context. In addition, due to the texts' antiquity, most wooden block prints and manuscripts have faded inks and minor stains on pages that further challenge Google OCR's performance. 

\begin{figure}[ht]
\centering
\includegraphics[width=0.75\linewidth]{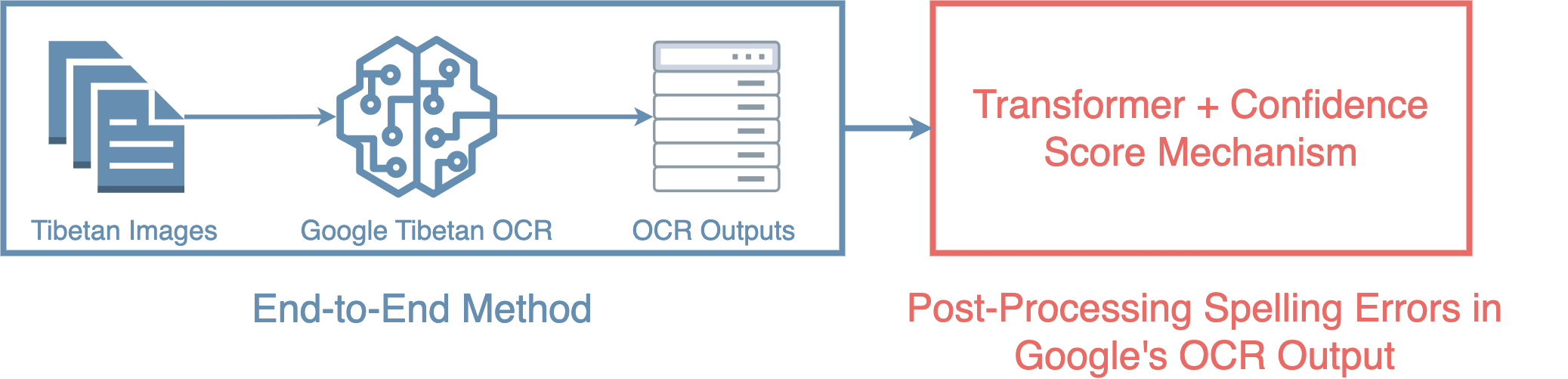}
\caption{Intended Usage: Unlike the common End-to-End method, which processes original raw images into texts, our Transformer + Confidence Score mechanism is designed for post-processing spelling errors in Google's OCR outputs.}
\label{fig:3b}
\end{figure}

Current popular spelling correction systems cannot correct Tibetan texts and are not designed to auto-correct OCR output. The OCR output contains a confidence score for each character it recognizes. In general, characters with higher than 80\% confidence scores have an overall accuracy close to 100\% based on human validation, and characters with low confidence scores, such as 30\%, require special attention. Recent NLP advances in Transformer architecture \cite{1}, which primarily consists of a multi-head self-attention mechanism stacked in combination with an encoder/decoder structure, have proven to be adept at correcting erroneous English text and performing context-based grammatical error correction on incorrect characters/words \cite{5,6,8}. Inspired by this idea, we built on these advancements by implementing an additional Confidence Score mechanism into the standard Transformer architecture to further improve Transformer's performance. Unlike the common End-to-End method, which processes original raw images into texts, our Transformer + Confidence Score mechanism is designed for post-processing spelling errors in Google's OCR outputs, as shown in Figure \ref{fig:3b}. Through the Confidence Score mechanism, the model is able to take in both OCR-ed noisy sentences and the confidence score corresponding to each token in the encoder, and outputs corrected sentences in the decoder. The paired training data has an initial 25\% error rate, and our best model reduced the error rate to 12.26\%. Our model can furthermore be adapted to all languages dealing with post-processing OCR outputs.

This paper largely consists of four sections: data, model architecture, training, and analysis. First, we paired our training data with both Google’s Tibetan OCR outputs and human-validated clean data. Then, we implemented a Confidence Score mechanism into the Transformer architecture to perform spelling correction tasks. According to the Loss and Character Error Rate metrics, our Transformer + Confidence Score mechanism architecture proves to be superior to the Transformer, LSTM-to-LSTM, and GRU-to-GRU architectures. Finally, to examine the robustness of our model, we analyzed erroneous tokens and visualized Attention and Self-Attention heatmaps. The initial error rate of Google’s Tibetan OCR outputs was 25\%, and our best model reduced this error rate to 12.26\%. Our model can furthermore be adapted to any language dealing with post-processing OCR-ed texts.

\section{Related Work}\label{se:2}

While recent advances in NLP have resulted in significant progress on real-world language-based tasks, there have been few attempts in applying these advances to the field of Tibetan studies. Powerful deep seq2seq architectures \cite{9}, such as the Transformer~\cite{1}, have demonstrated great improvement in translation tasks. The Transformer relies heavily on Attention mechanisms~\cite{4} with stacked encoders and decoders. Each of these components consists of multi-headed Self-Attention layers followed by position-wise feed-forward layers, a normalization layer, and other residual connections. Notably, the decoder blocks attend to and are connected to encoder outputs via these Self-Attention and feed-forward layers. These complex Attention mechanisms have shown rich language understanding compared to more complex recurrent or convolutional neural networks on English-to-German and English-to-French translation \cite{1}. 

Many recent neural Grammatical Error Correction (GEC) models are trained on this type of Transformer architecture \cite{5,6,8}. As a whole, our task of correcting the spelling errors in the OCR-ed Tibetan corpus is very similar to the GEC task. In particular, Shahgir et al. \cite{shahgir2023bangla} employed a Text-to-Text Transfer Transformer (T5) Language Model to detect grammatical errors in Bangla, demonstrating that the T5 model can achieve a low Levenshtein Distance. The authors also suggested that their approach can be adapted for other languages. Choe et al. \cite{3} built a GEC system that was trained on artificially generated data, with researchers generating an erroneous corpus using a realistic noising function. Given low resource machine translation and a lack of data, the researchers randomly inserted/replaced/removed tokens or swapped neighboring words with a uniform distribution. This noising function was shown to be effective at replicating real-world conditions and, ultimately, this corpus was then used to train the Transformer architecture. 

Previous work \cite{8} implemented a Copy mechanism into the Transformer architecture to perform GEC tasks, demonstrating that about 80\% $\sim$ 97\% tokens are directly copied from the source sentence. Similar to the Attention~\cite{4}, this Copy mechanism allows the model to identify which word should be simply copied in the decoder while correcting a sentence, and has proven to be effective for text summarization~\cite{see2017get, gu2016incorporating} and semantic parsing~\cite{jia2016data}. Zhao et al.~\cite{8} also implemented the Copy-Augmented mechanism with a denoising auto-encoder to improve GEC performance. In our task of correcting errors in the OCR-ed Tibetan corpus, the OCR system outputs a confidence score for each character. We do not need to incorporate this Copy or Copy-Augmented mechanism because we already have the confidence score that represents the trustworthiness of a given character. 

Furthermore, a compression algorithm \cite{7} was adapted from the byte-pair encoding (BPE) to perform word segmentation, which allows for a fixed-size vocabulary of variable-length character sequences. This subword unit segmentation method introduced by Sennrich et al. \cite{7} is particularly helpful in implementing a GEC task model, because it grants the model the flexibility of changing a subunit of a word. Ingólfsdóttir et al.'s work~\cite{ingolfsdottir2023byte} also show that BPE can enable higher correction quality than a subword approach for GEC, as it can handle not only the simple spelling errors, but also for more complex semantic, stylistic and grammatical issues.

In recent years, OCR has evolved with the end-to-end solution of deep learning, especially transformer-based architectures~\cite{subramani2020survey, li2021trocr}. People also developed masking system to detect and mask non-textual artifacts in the noisy images before feeding them into the OCR system to improve OCR results. The above papers all start with image input. However, in this paper, we do not propose methods to process images but instead focus on post-processing the OCR output text similar to Rakshit et al.~\cite{rakshit2023novel}. Both of our and Rakshit et al.’s approaches work on post-processing errors in OCR-ed text, but Rakshit et al.’s study focuses on English-language data. This post-processing is particularly valuable when we are unable to alter the OCR system itself or when there is a scarcity of image-text pairs for training but an abundance of unlabeled text data.

\section{Data}\label{se:3}

In collaboration with BDRC, we received both human-corrected data and OCR-ed noisy data for our project. The human-corrected data has been meticulously proofread by field experts and contains minimal errors. Therefore, we used it as the ground truth for training our model. The raw output from Google’s Tibetan OCR system, which is the noisy data, is provided in JSON format. The JSON files include: 1) the x and y coordinates of each recognized syllable, 2) the recognized syllable itself, and 3) the confidence score for the recognized syllable. We extracted the recognized syllables and their corresponding confidence scores from the JSON files and aligned them with the human-corrected data. Table. \ref{tab:1} presents a sample of our training dataset. 

\begin{table}[ht]
\caption{Sample Training Data. The column “Noisy data from OCR” contains syllables recognized by Google's Tibetan OCR system. The middle column “Confidence score from OCR” contains lists of scores ranging from 0 - 1.0, which indicate the OCR system's confidence in the recognition of each syllable. Thus, the syllables in the left column and the confidence scores in the middle column are perfectly paired by index. The third column, “Clean data,” contains the human-corrected data. The human-corrected data is paired with the noisy data at the sentence level.}
\label{tab:1}
\begin{tabular}{rrrr}
\toprule
 & Noisy data from OCR & Confidence score from OCR & Clean data\\
\midrule
0 & \includegraphics[scale=0.1]{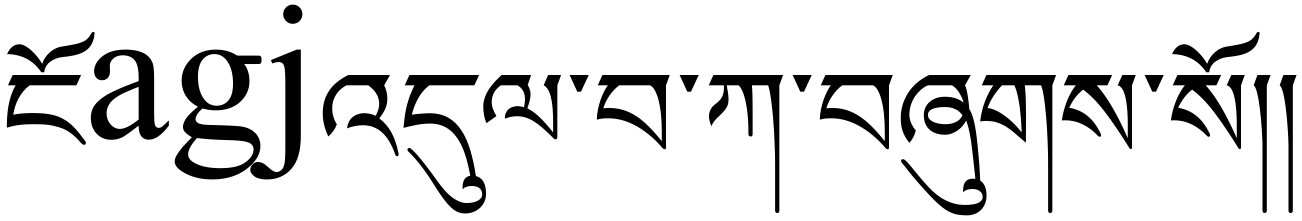} & [0.32, 0.32, 0.14, 0.7, 0.24, 0.99, 1.0, 1.0, ... & \includegraphics[scale=0.1]{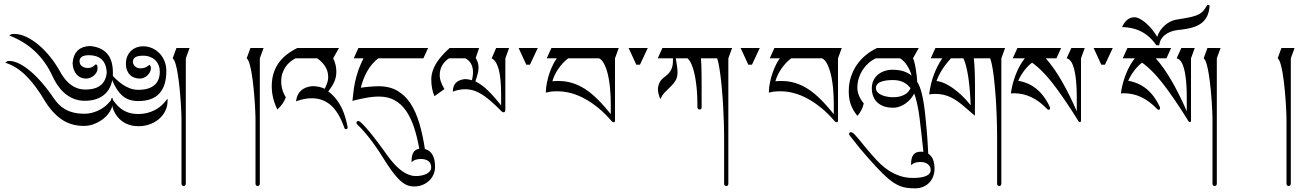}\\
\multirow{2.5}*{1} & \multirow{2.5}*{\includegraphics[scale=0.1]{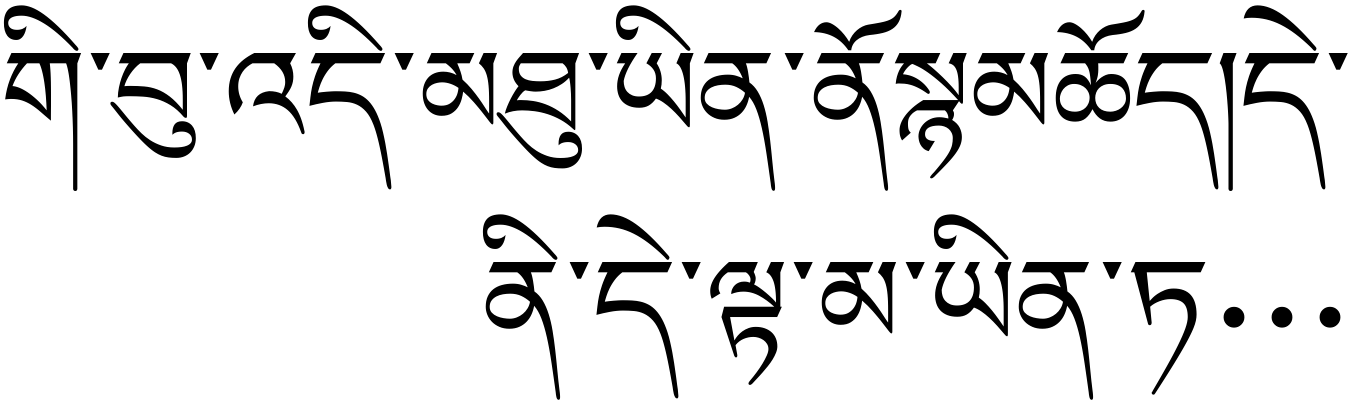}} &  \multirow{2.5}*{[0.99, 0.99, 0.97, 0.65, 0.65, 0.9, 0.87, 0.99...} & \multirow{2.5}*{\includegraphics[scale=0.1]{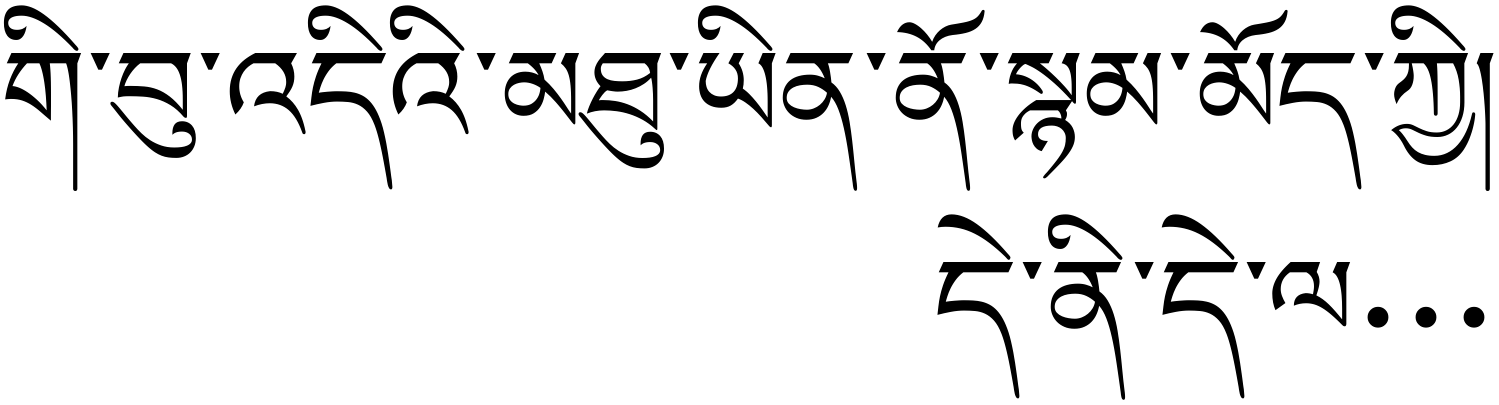}}\\
&&&\\
\multirow{2.5}*{2} & \multirow{2.5}*{\includegraphics[scale=0.1]{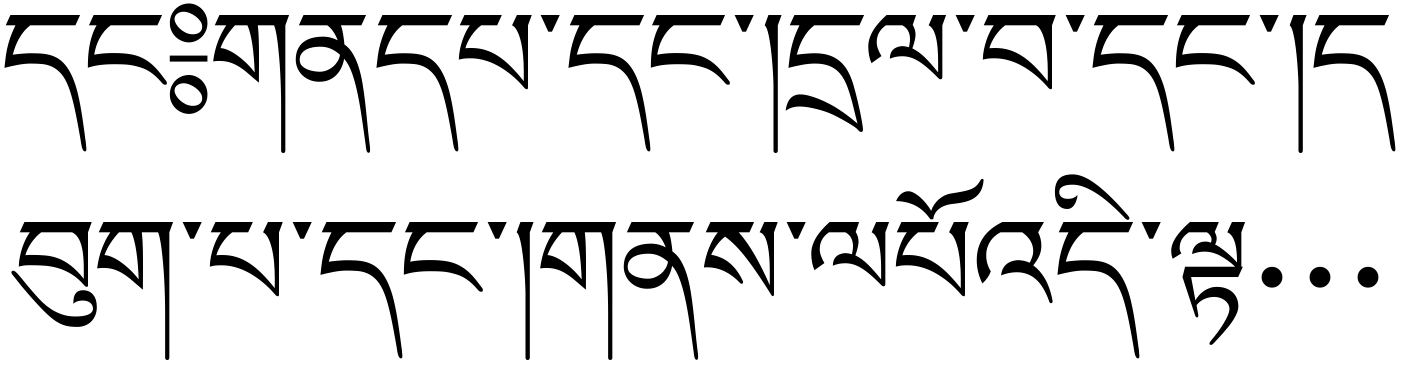}} & \multirow{2.5}*{[0.99, 0.87, 0.07, 0.26, 0.28, 0.6, 0.65, 0.88...} & \multirow{2.5}*{\includegraphics[scale=0.1]{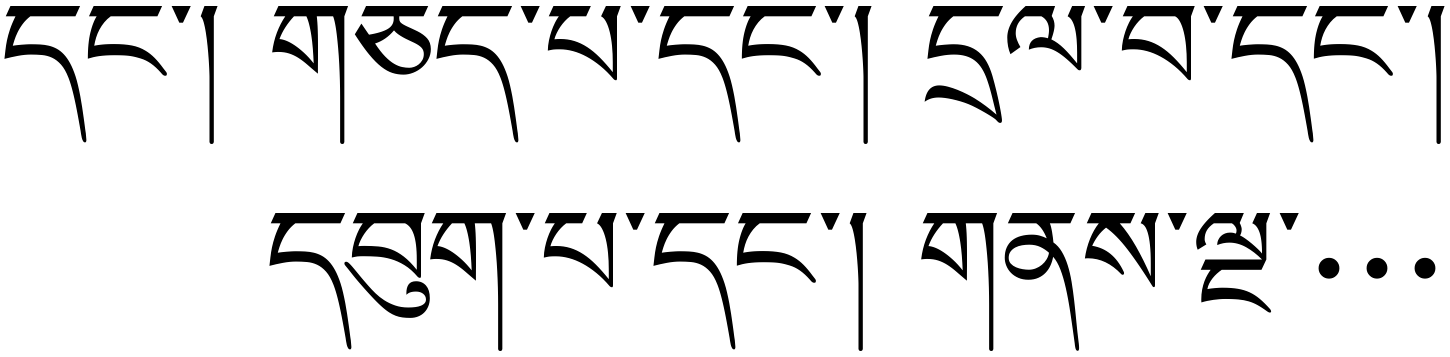}}\\
&&&\\
\bottomrule
\end{tabular}
\end{table}



\section{Model Architecture}\label{se:4}

In our task of correcting Tibetan sentences, we choose a seq2seq architecture~\cite{9, 1} to translate the raw OCR-ed noisy texts as input sequences $X$ into cleaned texts as output sequences $Y$. A seq2seq architecture contains an encoder and a decoder. The encoder maps an input sequence of $x=\left[x_1, x_2, \ldots, x_n\right]$ to a latent representations of $z=\left[z_1, z_2, \ldots, z_n\right]$, where:
$$
z = encoder(x)
$$
Given $z$, the decoder generates a probability distribution over output sequences $y=\left[y_1, y_2, \ldots, y_n\right]$ one at a time: 
$$
\log p(y \mid x)=\sum_{i=1}^y \log p\left(y_i \mid y_{<i}, z, x\right)
$$

Our baseline model is the Transformer~\cite{1}, which follows this architecture with multi-headed self-attention and position-wise feed-forward layers in both the encoder and decoder, and the decoder uses attention to attend to the encoder. The Transformer's encoder takes in the OCR-ed noisy sentence and its decoder outputs the corrected sentence, then the model computes the loss and updates the weights by comparing its predictions with the ground truth sentence. 

\begin{figure}[ht]
\centering
\includegraphics[width=0.75\linewidth]{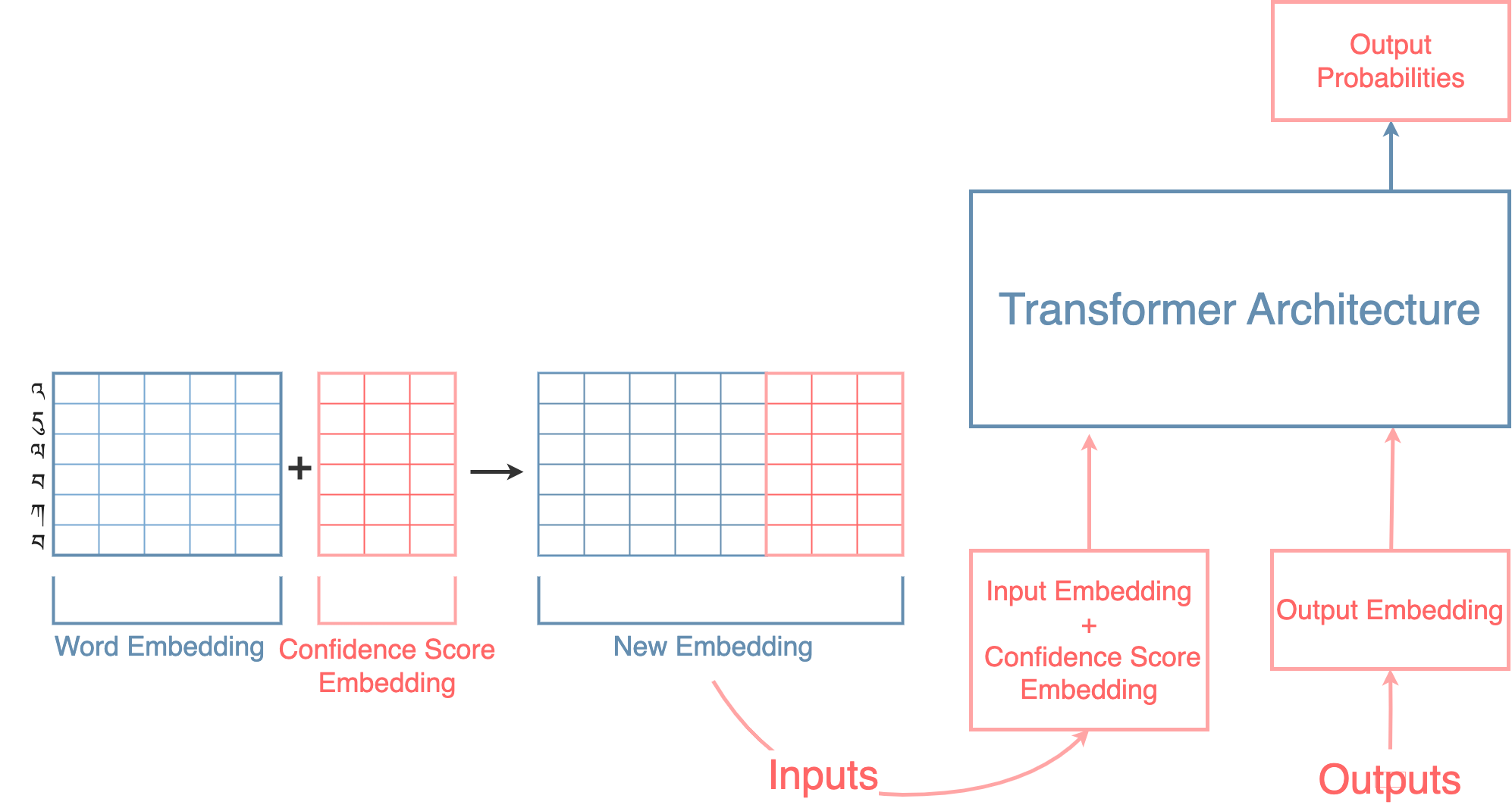}
\caption{Model Architecture. The Confidence Score mechanism is incorporated into the Transformer architecture, designed to integrate OCR system confidence scores. This is achieved by augmenting the standard input embeddings with additional confidence score embeddings.}
\label{fig:3}
\end{figure}

Built on the Transformer architecture, we also implemented an additional mechanism to incorporate the confidence scores from the OCR system. As Figure. \ref{fig:3} shows, we augment the inputs by concatenating input embeddings with confidence score embeddings. The weights of confidence score embeddings are not randomly initialized. Since the confidence scores range between 0-1, we manually assigned weights between 0-1 to facilitate fast learning during training. For example, if a given character has a confidence score of 0.23, its confidence score embedding weights are initialized as 0.23. 

This Confidence Score mechanism was derived from the implementation of a Copy mechanism \cite{8} into their Transformer architecture to perform GEC tasks, in which the Copy mechanism learns a set of weights to inform the model whether a given token should be simply copied to the decoder or not. Recognizing the importance of informing the model about whether a particular token needs to be changed or not, we chose to implement a Confidence Score mechanism in-place of the Copy mechanism since it fits better for our task of post-processing OCR-ed texts; notably, this deviation does not affect weighting structure of the model given that the confidence scores are equivalent to the weights learnt from the Copy mechanism. Overall, the confidence score embedding was chosen to be able to capture the distributions of low-confident tokens and the interdependencies between low-confident tokens and their high-confident neighbors during training, and thus improve the model's performance. We will do a thorough analysis on the Confidence Score mechanism and visualize the Self-Attention in the analysis section.

\section{Training, Evaluation, Results}\label{se:5}
\subsection{Evaluation Metrics}\label{se:5.1}

To understand the efficacy of the NLP-corrected model output, we use both Loss, and Character Error Rate (CER) as evaluation metrics to examine our model's performance. 

\vspace{1em}\noindent\emph{Loss:}

As an evaluation metric and a value for the loss function, we use the Kullback-Leibler (KL) divergence because it provides a distance-based measurement for continuous distributions and allows us to conduct discrete sampling across a continuous output distribution. Ultimately, this means that the loss function allows us to compare the model's predicted probability distribution to the label-smoothed target probability distribution, and measures relative entropy and randomness in continuous sequences. 

\vspace{1em}\noindent\emph{KL divergence equation:}
$$
D_{K L}(P \| Q)=\sum_{x \rightarrow X} P(x) \log \left(\frac{P(x)}{Q(x)}\right)
$$
Where $Q$ is a prior probability distribution, $P$ is posterior probability distribution. $D_{K L}(P \| Q)$ is the relative entropy from $Q$ to $P$ or the measure of gained information by revising one's beliefs from $P$ given $Q$.

\vspace{1em}\noindent\emph{Character Error Rate:}

To quantify the similarity between the NLP-corrected model output and the ground-truth string, we primarily use Levenshtein distance or edit distance, which allows us to capture similarity by calculating the minimum number of single-character edits (insertions, deletions or substitutions) required to change the ground truth text to the model output which helps us build a better picture of how close the model's output is from the ground truth. For this project, we specifically use the Levenshtein distance to calculate the ratio of matched string blocks by character which gives an aggregate character error rate (CER).
\begin{itemize}
\item[]\hspace{-1em} CER $= \frac{S+D+I}{N}$
\begin{itemize}
\item[-] $S$ is the number of substitutions
\item[-] $D$ is the number of deletions
\item[-] $I$ is the number of insertions
\item[-] $N$ is the number of characters in the reference
\end{itemize}
\end{itemize}

Effectively, CER serves as a ratio between the total number of incorrect characters divided by the total number of correct characters. The larger the CER, the greater the difference between the two target texts.

\subsection{Training}\label{se:5.2}

\subsubsection{Tokenizer Vocabulary Size}\label{se:5.2.1}

The Transformer $+$ Confidence Score mechanism architecture concatenates the input embedding with the confidence score embedding, so we need to finetune both the tokenizer vocabulary size for input embedding and the confidence score vocabulary size for confidence score embedding. We first trained a BPE tokenizer for input embedding using the noisy sentences to include the erroneous tokens, so that the model can easily recognize the erroneous tokens during training. We experimented with various vocabulary sizes of 100, 300, 500, 1000, 2000 to find out the optimal parameter. For a BPE tokenizer, a small vocabulary size, such as 30, is equivalent to a character-level tokenization which only contains the most fundamental alphabets as vocabularies. On the other hand, a large vocabulary size, say 2000, is close to a word-level tokenizer that includes much more common consecutive sequences of alphabets, such as ``ed'', ``ing'' and ``n't'' in English. Since our noisy data contains confidence scores for each character, it is intuitive to use character-level tokenization and choose a small vocabulary size to fully utilize this valuable information, but such small vocabulary size might limit the model's capacity to capture meaningful semantic relationships. On the other hand, a large vocabulary size might require a much larger training set and computational power to fully populate the embedding space of each token. As Table. \ref{tab:2} shows, given other parameters set the same, the best BPE tokenizer vocabulary sizes are 300 and 500. The model tends to overfit the training set with vocabulary size of 100, while underfit with vocabulary size of 1000 and 2000. This may be due to the fact that the most common combinations of syllables in our Tibetan language corpus were around 300 to 500, resulting in the most efficient and dense representation in the model.
\begin{table}[ht]
\caption{Training Loss and CER with Respect to Different BPE Tokenizer Vocabulary Sizes: In this scenario, we used a confidence score vocabulary size of 101 for all cases. We found that a BPE tokenizer vocabulary size of 500 was the best on both metrics of Loss and CER.}
\label{tab:2}
\begin{tabular}{lp{2cm}p{2.4cm}ll}
\toprule
 & BPE tokenizer vocab size & Confidence score vocab size & \multirow{2}*{Loss} & \multirow{2}*{CER}\\
\midrule
Transformer $+$ Confidence Score Mechanism & 100 & 101 & 0.12
  & 0.13\\
Transformer $+$ Confidence Score Mechanism & 300 & 101 & 0.12 & 0.13\\
Transformer $+$ Confidence Score Mechanism & {\bf 500} & {\bf 101} & {\bf 0.12} & {\bf 0.12}\\
Transformer $+$ Confidence Score Mechanism & 1000 & 101 & 0.20 & 0.37\\
Transformer $+$ Confidence Score Mechanism & 2000 & 101 & 0.25 & 0.46\\
\bottomrule
\end{tabular}
\end{table}

\subsubsection{Confidence Score Vocabulary Size}\label{se:5.2.2}

Next, we tuned the confidence score vocabulary size and experimented with three design choices of vocabulary sizes 101, 5 and 2. The OCR confidence scores are floats from 0 to 1 $(0.0, 0.10, 0.11 \ldots 1)$, so we started with a vocabulary size of 101. This vocabulary size allows the model to learn the difference between scores of 0.89 and 0.90. However, this vocabulary size required a large dataset to populate every score from 0 to 1. We then considered a vocabulary size of 5, where floats between 0-0.2 are mapped to 1, floats between 0.2-0.4 to 2, and so on. This method largely reduced the vocabulary size to produce less trainable weights and thus shortened the training time. However, doing so might have potentially oversimplified the problem as the model could no longer learn more nuanced differences between scores of 0.80 and 0.90. Lastly, we implemented the simplest scheme of using a vocabulary size of 2, where scores between 0 and 0.8 are mapped to 0 and scores bigger than 0.8 are mapped to 1, representing a binary representation of not-confident vs. confident. We chose the threshold as 0.8 because alphabets with higher 0.8 scores are generally all correct based on human validation. We expected the model to distinguish non-confident vs. confident alphabets, but experimentally, the model did not demonstrate apparent improvement over the original Transformer architecture. 

Table. \ref{tab:3} shows the training Loss and CER with respect to different confidence score vocabulary sizes. Based on the Loss and CER, the vocabulary size of 101 is significantly better than that of 5 and 2. This indicates that 1) the Confidence Score mechanism is effective in providing meaningful information to the model, 2) the nuanced signals in confidence scores ranging between 0-1.0 provided by the OCR system are important, and 3) the vocabulary sizes of 5 and 2 are not expressive enough in capturing the general confidence-trend in these scores.

Finally, we chose a maximum sentence length of 512 and batch size of 2048 as they are the optimal parameters presented in the original Transformer paper. 
\begin{table}[ht]
\caption{Training Loss and CER with respect to different Confidence Score Vocabulary Sizes. In this scenario, we used BPE tokenizer vocabulary size of 300 for three cases of vocabulary size = 101, 5, and 2. Our analysis indicated that a vocabulary size of 101 is optimal for the model to effectively capture nuances in numerical values. In contrast, vocabulary sizes of 5 and 2 lose meaningful information in the confidence scores provided by the OCR system.}
\label{tab:3}
\setlength{\tabcolsep}{1.2mm}{
\begin{tabular}{lp{2cm}p{2.3cm}ll}
\toprule
 & BPE tokenizer vocab size & Confidence score vocab size & \multirow{2}*{Loss} & \multirow{2}*{CER}\\
\midrule
Transformer $+$ Confidence Score Mechanism & {\bf 300} & {\bf 101} & {\bf 0.1228} & {\bf 0.1265}\\
Transformer $+$ Confidence Score Mechanism & 300 & 5 & 0.1485 & 0.1753\\
Transformer $+$ Confidence Score Mechanism & 300 & 2 & 0.1501
  & 0.1780\\
\bottomrule
\end{tabular}}
\end{table}


\subsection{Results}\label{se:5.3}

Overall, our Transformer $+$ Confidence Score mechanism architecture outperforms the rest of the architectures (Table. \ref{tab:5}). Compared to the Transformer architecture, our model improved the Loss by 23.26\% and CER by 30.66\%. These results suggest that the Confidence Score mechanism is able to capture meaningful information from the confidence scores provided by Google’s OCR system. As a corpus-based metric, a CER of 0.1226 suggests an in-depth ability to understand and predict individual words and phrases. Compared to LSTM-to-LSTM or GRU-to-GRU, which both suffer from a significantly higher CER, the Transformer $+$ Confidence Score mechanism is able to not only boost performance across all three categories of interest but also exhibit a more well-rounded, complete understanding of Tibetan semantics and vocabulary.
\begin{table}[ht]
\caption{Model Performance Comparison. The Transformer + Confidence Score mechanism outperformed the other three model architectures. In particular, compared to the Transformer architecture, it improved the Loss and CER by 23.26\% and 30.66\%, respectively.}
\label{tab:5}
\setlength{\tabcolsep}{3mm}{
\begin{tabular}{lll}
\toprule
 & Loss & CER\\
\midrule
Transformer & 0.1509 & 0.1768\\
{\bf Transformer $+$ Confidence Score mechanism} & {\bf 0.1158} & {\bf 0.1226}\\
LSTM-2-LSTM & 0.2419 & 0.4753\\
GRU-2-GRU & 0.2584 & 0.4831\\
\bottomrule
\end{tabular}}
\end{table}

\section{Analysis and Discussion}\label{se:6}

Having demonstrated the efficacy of the Transformer $+$ Confidence Score mechanism, we conduct an in-depth analysis of key sub-modules within the model architecture to better understand its strengths and weaknesses. In evaluating the model, we specifically analyzed erroneous tokens to identify the strengths and weaknesses of the model's learned representations, visualized Attention and Self-Attention heatmaps within the Transformer, and explored important neurons in our model. 

\subsection{Visualizing Erroneous Tokens}\label{se:6.1}

\begin{figure}[!hb]
\centering
\subfigure[]{
\label{fig:4a}
\includegraphics[width=0.35\linewidth]{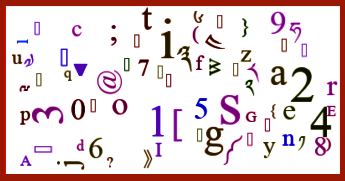}}
\subfigure[]{
\label{fig:4b}
\includegraphics[width=0.353\linewidth]{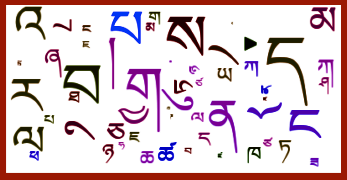}}
\caption{The figure on the left shows tokens that the model generally succeeds in correcting. These tokens are so distinctly different from Tibetan alphabets, such as ``y'', ``7'', and ``@'' that the model can easily identify and edit them. The figure on the right shows tokens that the model generally fails to correct. Tokens such as ``\includegraphics[scale=0.025]{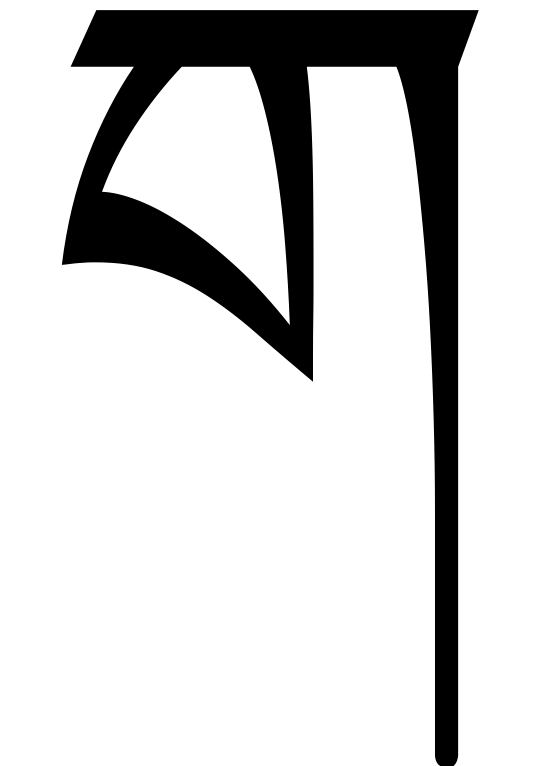}'', ``\includegraphics[scale=0.025]{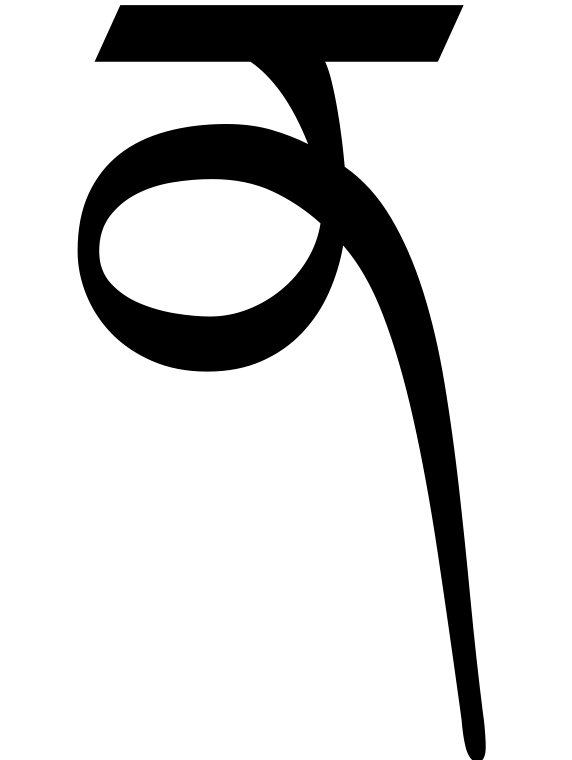}'', and ``\includegraphics[scale=0.02]{samples/b.png}'' are among the most common Tibetan alphabets, which can combine with numerous other Tibetan syllables to form meaningful words.}

\label{fig:4}
\end{figure}

To identify the recurring patterns and systematic errors in the model, we visualized and analyzed: 1) tokens that the model generally succeeds in correcting (Figure. \ref{fig:4a}); 2) tokens that the model generally fails to correct (Figure. \ref{fig:4b}). 

Figure. \ref{fig:4a} shows tokens that the model generally \emph{succeeds} to correct, among which are non-Tibetan alphabets and symbols, such as ``y'', ``7'', and ``@''. These tokens are so distinct from Tibetan alphabets that the model can easily pick up and edit these tokens. On the other hand, the model \emph{cannot} easily recognize and successfully edit Tibetan alphabets, such as ``\includegraphics[scale=0.025]{samples/k.png}'', ``\includegraphics[scale=0.025]{samples/l.png}'', and ``\includegraphics[scale=0.02]{samples/b.png}'' shown in Figure. \ref{fig:4b}. There are several factors that contribute to this type of error. First, ``\includegraphics[scale=0.025]{samples/k.png}'', ``\includegraphics[scale=0.025]{samples/l.png}'', and ``\includegraphics[scale=0.02]{samples/b.png}'' are the most common Tibetan alphabets that they can combine with many other alphabets to make a meaningful word. In a test sentence, the model fails to correct ``\includegraphics[scale=0.025]{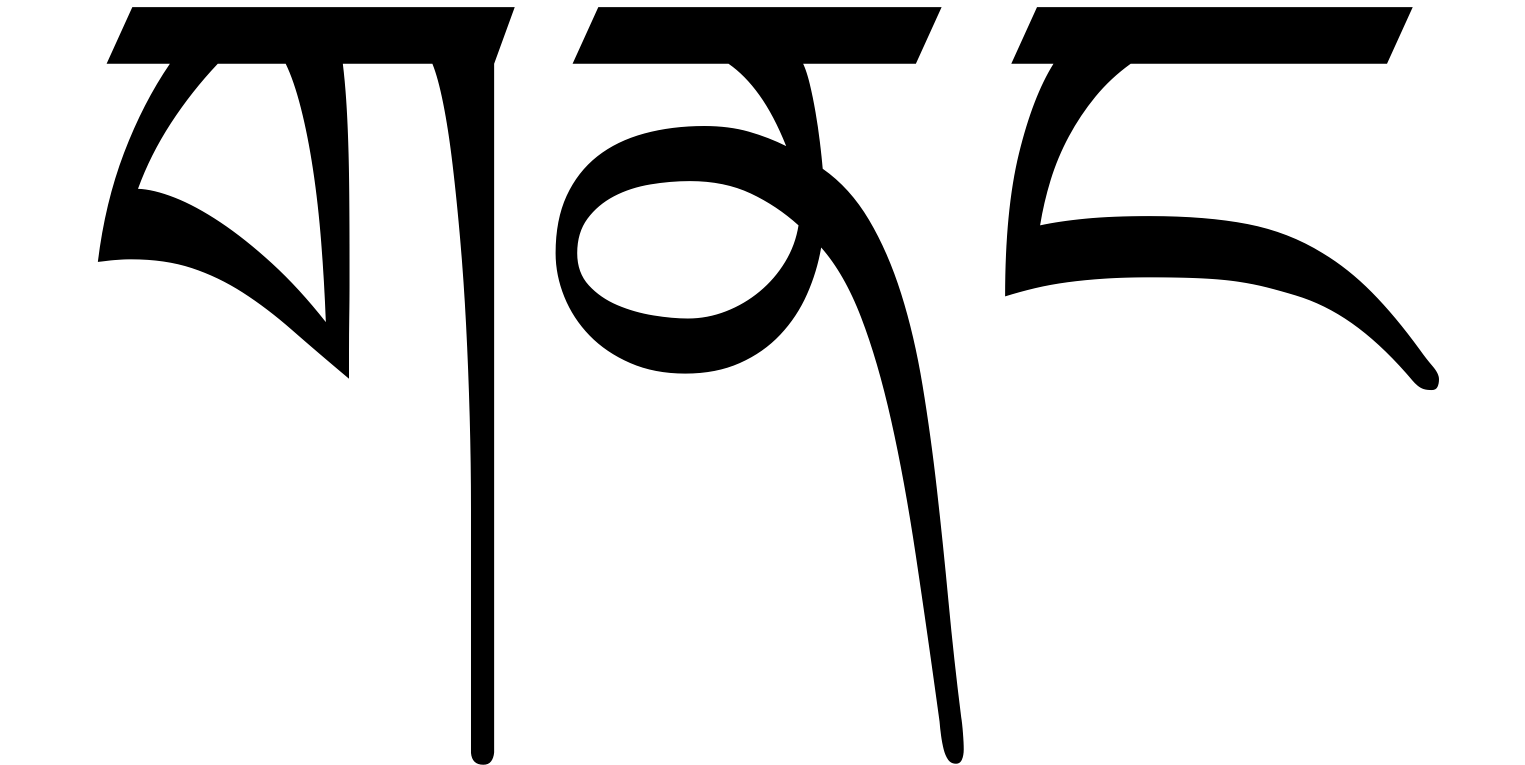}'' to ``\includegraphics[scale=0.025]{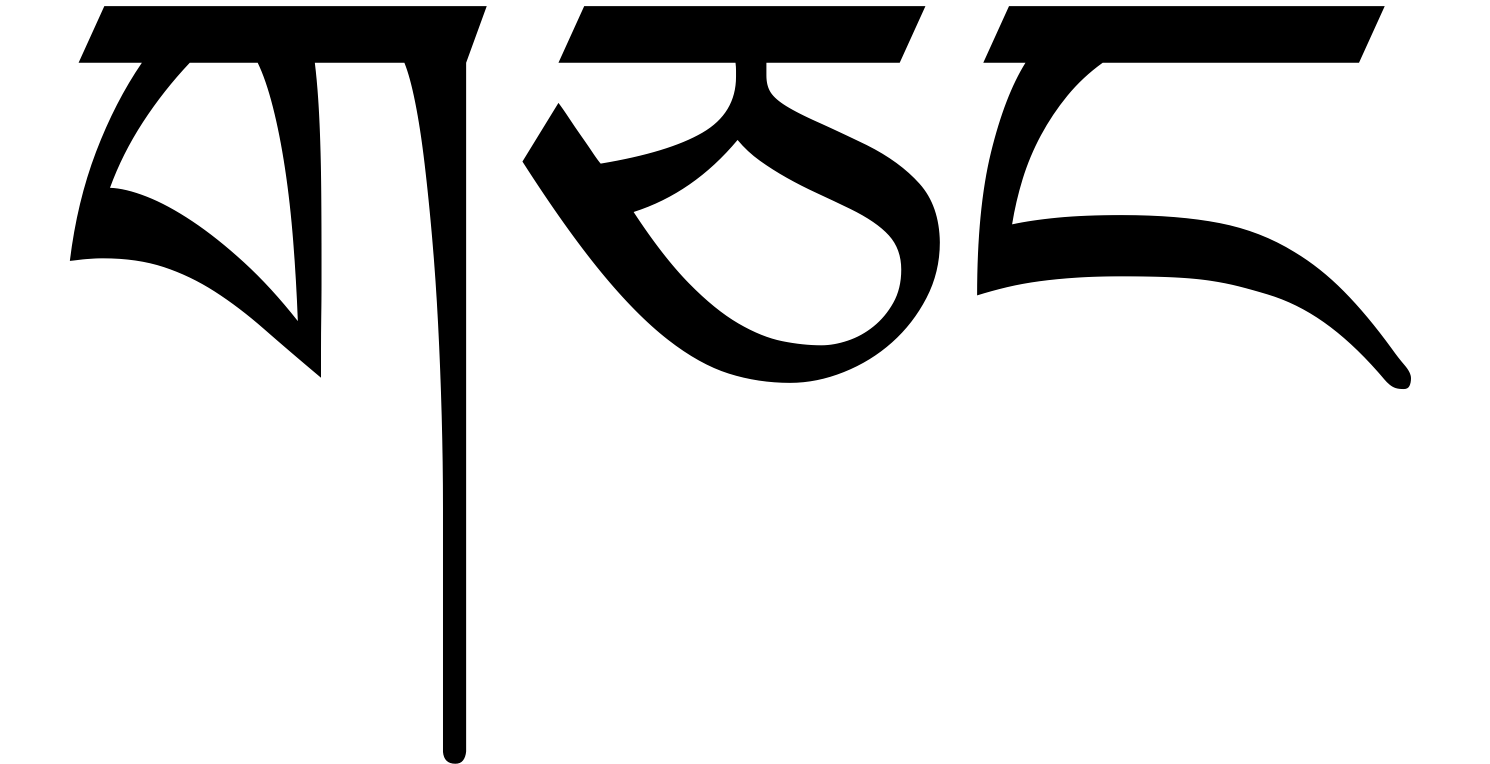}'', in which it fails to correct ``\includegraphics[scale=0.025]{samples/l.png}'' to ``\includegraphics[scale=0.025]{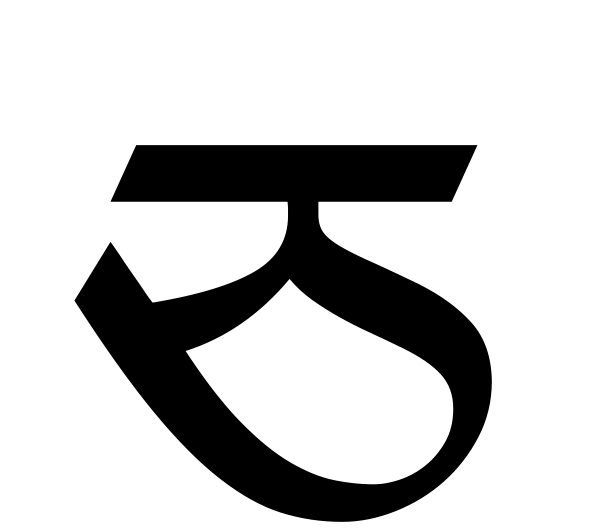}'' because both spellings are legitimate. The former means ``sanction'' and the latter means ``cleverness''. The former is a wrong word because it does not fit in the sentence's context. Second, compared to the general BPE tokenizer vocabulary size of 3000-5000 for an English corpus, our best model has a tokenizer vocabulary size of 500. Such a small vocabulary size results in that our model cannot capture good contextual meaning between words. The Attention analysis in the following section also demonstrates the same weakness in our model.  

\subsection{Visualizing Attention}\label{se:6.2}

\begin{figure}[ht]
\centering
\subfigure[]{
\label{fig:5a}
\includegraphics[width=0.35\linewidth]{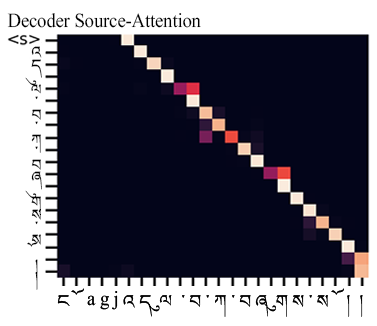}}
\subfigure[]{
\label{fig:5b}
\includegraphics[width=0.26\linewidth]{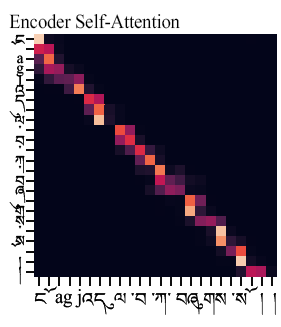}}
\subfigure[]{
\label{fig:5c}
\includegraphics[width=0.275\linewidth]{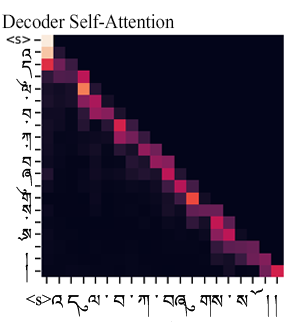}}
\caption{Attention Heatmaps. The attention heatmaps are generated using the third attention layer, averaged across four heads in both encoder and decoder. Dark squares indicate low attention weights and correlations between tokens, while bright squares indicate high attention and correlations. Subfigures (a), (b), and (c) show the Source-Attention in the decoder, and Self-Attention in the encoder and the decoder, respectively.}
\label{fig:5}
\end{figure}

An Attention heatmap is a visual representation that shows the attention weights between two tokens. It enables us to see whether the model is attending to the correct tokens while generating the output. The Transformer architecture contains two types of Attention: Self-Attention and Source-Attention. The Transformer's encoder contains only the Self-Attention which allows the input to interact with itself, while the decoder contains both Self-Attention and Source-Attention, in which the Self-Attention allows the output to interact with itself, and the Source-Attention allows the output to look at the input. There are three attention layers in both the encoder and the decoder, and each attention layer contains four attention heads. The attention heatmaps (Figure \ref{fig:5}) are generated using the third attention layer in both the encoder and decoder, and averaged across four attention heads. 

In Figure. \ref{fig:5}, the dark squares represent low attention weights between the tokens on the x-axis and the y-axis, indicating low correlations between these tokens, whereas the bright squares represent high attention weights and high correlations. Figure. \ref{fig:5a} shows the Source-Attention heatmap from the decoder. ``\includegraphics[scale=0.025]{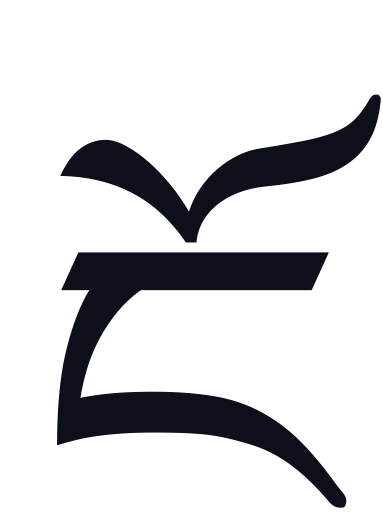}agj'' on the far left of x-axis are the noises from OCR that should be eliminated in the output. We can see that the decoder has low attention weights on ``\includegraphics[scale=0.025]{samples/p.png}agj'', but has high attention weights on the parallel units of the sentence (``\includegraphics[scale=0.06]{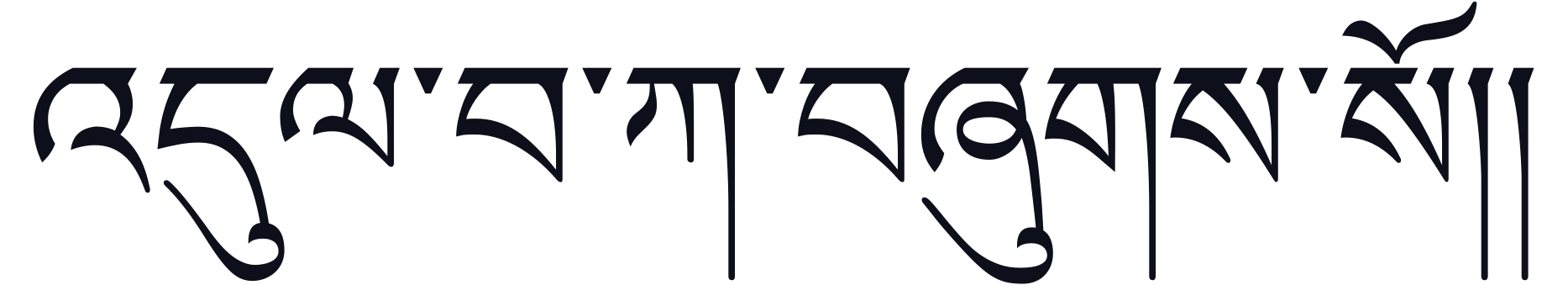}''), indicating the model is looking at the correct characters while generating output. 

Figure. \ref{fig:5b} and \ref{fig:5c} show Self-Attention weights in the encoder and decoder, respectively. One phenomenon the Self-Attention figures tended to exhibit is a strong awareness of its nearest neighbors, effectively meaning most of its attention is directed towards 2-3 neighbors to the left and right of the current token. Self-attention directed towards close neighbors is extremely important for correcting spelling errors within the OCR-ed text and helps to correct erroneous characters within a given word. Take an example in English for readability: by looking at close neighbors, the model is able to correct ``havard'' to ``harvard'' by inserting a ``r'' in the middle and it can similarly handle the challenge of  deleting textual noises such as random numbers, characters from the non-target language, or other out-of-place symbols from a given phrase.

However, given that these self-attention layers concentrated on close neighbors, farther neighbors received less attention regardless of whether they were related to the current token or not. Without attention on these farther neighbors, the model is unable to correct semantic errors that result from a lack of understanding about word context or grammatical context clues. Take the following pseudo-example for a hypothetical model input and output (For readability, this example is presented in English but the same sentiment exists within the model's Tibetan output.):

\vspace{1em}\underline{Ground truth:}

\emph{michael is playing with his \underline{kite} in the playground with friends}

\vspace{1em}\underline{Noisy OCR result:}

\emph{michael is playing with his \underline{kife} in the playground with friends}

\vspace{1em}\underline{Model correction:}

\emph{michael is playing with his \underline{knife} in the playground with friends}
\begin{figure}[ht]
\centering
\includegraphics[width=0.55\linewidth]{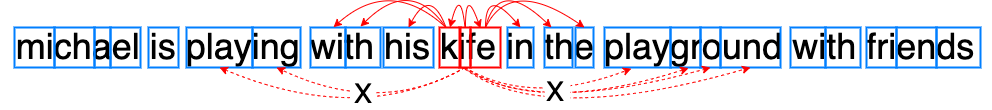}
\caption{An example of the model's short-sightedness, where ``kife'' is incorrectly corrected to ``knife'' instead of ``kite'' because it focuses on the nearest 2-3 neighbors. Note: This example is presented in English for readability, but a similar issue occurs in the model’s Tibetan output.}
\label{fig:6}
\end{figure}

In this example, we find that the model has corrected ``kife'' to ``knife'' instead of ``kite'' (Figure. \ref{fig:6}).  Given the model's short-sightedness, this makes sense given that ``knife'' is a more common word and given the fact that the model is unable to look further at context clues like ``playing'', ``playground'', or ``friends'' which could have helped to disambiguate word choice in this situation. 

One potential solution that can help address the model's current Self-Attention short-sightedness would be to increase BPE tokenizer size to a range between 2000 - 3000 like a general English Transformer model (the current model has a tokenizer size of 500) to allow the model to capture deeper semantic meaning. In the context of this project, limited computing resources means that this solution is left as an exercise for future research. 

\section{Conclusion}\label{se:7}

In conclusion, our Transformer $+$ Confidence Score model outperformed the Transformer, LSTM-to-LSTM, and GRU-to-GRU models. The model is robust in correcting spelling errors in Tibetan OCR-ed texts, but it can suffer. The model is robust in correcting spelling errors in Tibetan OCR-ed texts, but it can suffer short-sightedness that it cannot correct words based on word context. Possible solutions include: 1) increase the BPE tokenizer vocabulary size to 2000 - 3000 like a general English Transformer model and increase training data to allow the model to capture latent semantic relationships, and 2) train a Tibetan BERT model and connect it to the end of this model's outputs to edit semantically inconsistent and ambiguous words. 

We recognize that this particular project focused specifically on NLP methods to post-process OCR outputs. There is still much work to be done in building robust transcription models by combining both computer vision and NLP algorithms. Future areas of research for optimizing transcription performance include building more robust computer vision algorithms to optimize/enhance scan qualities and training denoising models to enhance image qualities to improve OCR accuracy.

\bibliographystyle{ACM-Reference-Format}
\bibliography{main.bib}




\end{document}